\documentclass{article} 
\usepackage[preprint]{colm2026_conference}

\usepackage{graphicx}
\usepackage{hyperref}       
\usepackage{url}            
\usepackage{booktabs}       
\usepackage{amsfonts}       
\usepackage{nicefrac}       
\usepackage{microtype}      
\usepackage{xcolor}         
\usepackage{amsmath}
\usepackage{amssymb}
\usepackage{mathtools}
\usepackage{amsthm}
\usepackage{float} 
\usepackage{bbm}
\usepackage{caption}
\usepackage{subfigure}
\usepackage{tabularx}
\usepackage{enumerate}
\usepackage{bm}
\RequirePackage{algorithm,algorithmic}
\usepackage{multirow}

\title{Meta-Learning at Scale for Large Language Models via \\ Low-Rank Amortized Bayesian Meta-Learning}

\author{Liyi Zhang \\
Department of Computer Science\\
Princeton University\\
\texttt{zhang.liyi@princeton.edu} \\
\And
Jake Snell \\
Department of Computer Science \\
Princeton University\\
\texttt{jsnell@princeton.edu} \\
\AND
Thomas L.~Griffiths \\
Department of Psychology and Computer Science\\
Princeton University \\
\texttt{tomg@princeton.edu}
}

\begin{document}

\maketitle

\begin{abstract}
Fine-tuning large language models (LLMs) with low-rank adaptation (LoRA) is a cost-effective way to incorporate information from a specific dataset. However, when a problem requires incorporating information from multiple datasets -- as in few shot learning -- generalization across datasets can be limited, driving up training costs. As a consequence, other approaches such as in-context learning are typically used in this setting. To address this challenge, we introduce an efficient method for adapting the weights of LLMs to multiple distributions, Amortized Bayesian Meta-Learning for LoRA (ABMLL). This method builds on amortized Bayesian meta-learning for smaller models, adapting this approach to LLMs by reframing where local and global variables are defined in LoRA and using a new hyperparameter to balance reconstruction accuracy and the fidelity of task-specific parameters to the global ones. ABMLL supports effective generalization across datasets and scales to large models such as \textsc{Llama3-8B} and \textsc{Qwen2-7B}, outperforming existing methods on the CrossFit and Unified-QA datasets in terms of both accuracy and expected calibration error. We show that meta-learning can also be combined with in-context learning, resulting in further improvements in both these datasets and legal and chemistry applications.

\end{abstract}

\section{Introduction}

Large language models (LLMs) handle a variety of tasks well by default \citep{radford2019language}. However, fine-tuning is often necessary to tailor LLMs to specific domains. While methods such as low-rank adaptation \citep[LoRA;][]{Hu2021LoRALA} fine-tune a pretrained LLM cost-effectively, the fine-tuned LLM is often limited to the domain it is trained on. Its performance may not improve in other domains and sometimes worsens because of catastrophic forgetting, which can erase existing capabilities  \citep{Lazaridou2021MindTG, luo2025empiricalstudycatastrophicforgetting}.

In-context learning (ICL) adapts models to new tasks by conditioning on examples from the task in the prompt \citep{min-etal-2022-metaicl, chen-etal-2022-meta}. However, prompt-based approaches have limited expressivity, and their gains may diminish as the complexity of in-context information rises \citep{dou2026clbenchbenchmarkcontextlearning}. In-weights meta-learning is another strategy for solving this problem, training models on a variety of tasks in a way that supports generalization across tasks \citep{Finn2017ModelAgnosticMF}. MAML-en-LLM \citep{mamlenllm}, adapts the Model-Agnostic Meta-Learning \citep[MAML;][]{Finn2017ModelAgnosticMF} framework to LLMs. However, this approach requires a large amount of computation and memory, using second-order gradient updates and saves a model for each task, making it challenging to apply at scale. 

We address this challenge by developing a meta-learning approach for LLMs that is practical to use at scale and can be combined with ICL. 
Reptile \citep{Nichol2018OnFM} is a meta-learning algorithm that has constant scaling with respect to the number of tasks, but it does not retain the full expressivity of MAML. We propose another scalable meta-learning solution based on Amortized Bayesian Meta-Learning \citep[ABML;][]{ravi2018amortized}. This method posits a generative model over model weights where task-specific weights are generated from global weights.  Conditional distributions over task-specific weights are shared across tasks, so computation and memory costs remain constant with respect to the number of tasks. This offers a path toward meta-learning for LLMs at scale, but introduces two challenges: specifying a generative model over LLM weight space, and making training stable when the scale of probabilities over model variables can overwhelm the data likelihood. 

\begin{figure}[t]
    \centering
    \includegraphics[width=1\linewidth]{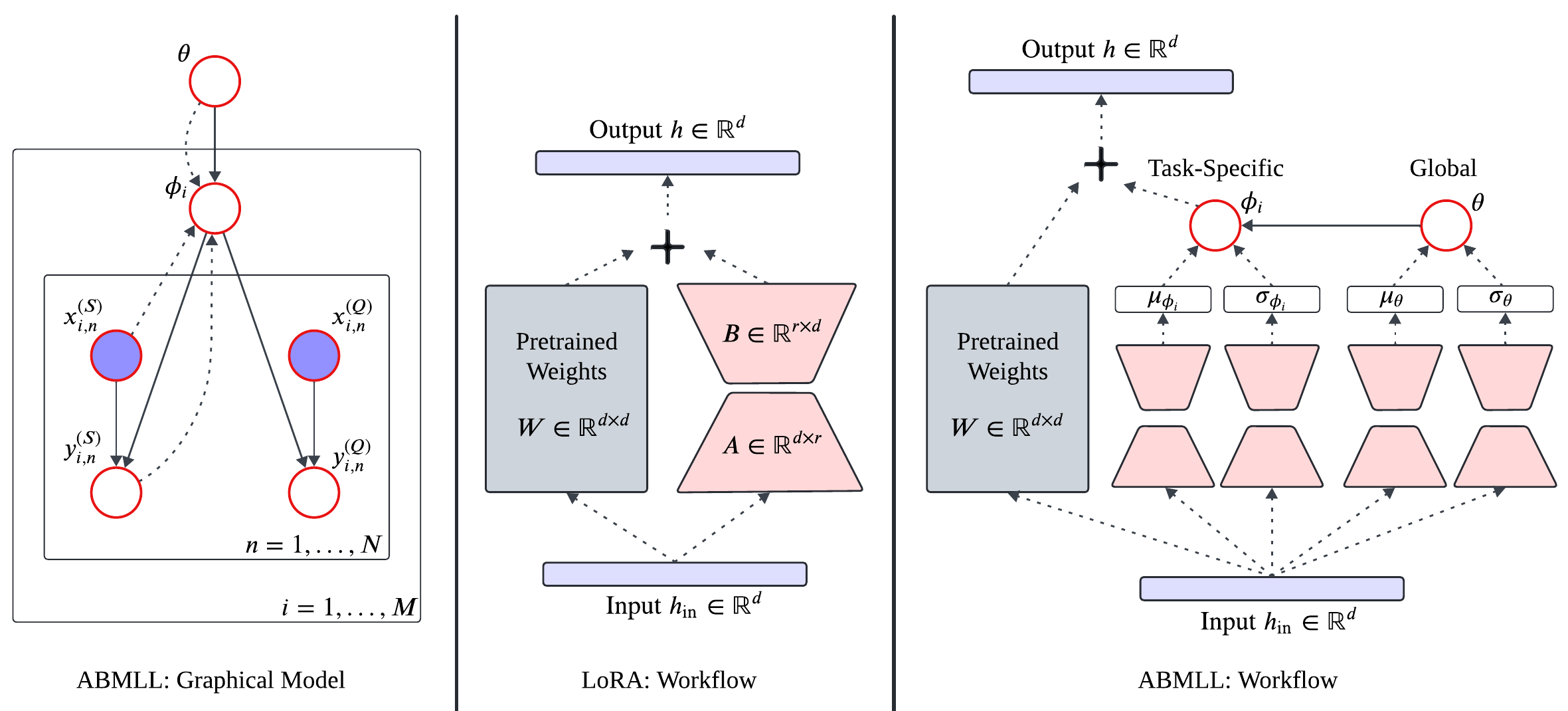}
    \caption{Illustrations of ABMLL and LoRA. There are $M$ tasks with $N$ datapoints each. $x$ is a prompt, $y$ is its output, and superscripts $S$ and $Q$ refer to the support set and the query set, which can be considered as train and test sets for individual tasks. Each solid arrow refers to a probabilistic relationship. On the graphical model shown on the left, a dashed arrow is a variational approximation; on the workflows shown to the right, a dashed arrow is an arithmetic operation.}
    \label{fig:intuitive-example}
\end{figure}

We present a solution to these two problems (see Figure \ref{fig:intuitive-example}). To define the generative model and efficiently characterize uncertainty, we use LoRA to express both model weights and their uncertainty, with mean and variance parameters computed as LoRA outputs. We introduce a new prior over global variables that accounts for the spread of pretrained parameters, along with a hyperparameter that balances reconstruction accuracy and the fidelity of task-specific parameters to global values.

Using Amortized Bayesian Meta-Learning for LoRA (ABMLL), we show that meta-learning can work effectively at the scale of \textsc{Llama3-8B} and \textsc{Qwen2-7B} \citep{grattafiori2024llama3herdmodels, yang2024qwen2technicalreport} with only a small memory increase over regular LoRA. On the standard benchmarks CrossFit and UnifiedQA, ABMLL improves over regular LoRA in both accuracy and expected calibration error, while performing competitively with other scalable meta-learning baselines on unseen tasks. We further show that meta-learning can be combined with ICL, yielding additional gains on these datasets as well as in legal and chemistry applications. Finally, we show that ABMLL improves robustness under pruning relative to regular fine-tuning and other meta-learning methods, suggesting the promise of a Bayesian approach for scalable and robust adaptation.

\section{Related Work}

\paragraph{Generalization methods in LLMs.} Extensive work has explored meta-learning as a method for improving generalization in machine learning systems, although these approaches were typically developed for models in the pre-LLM era \citep{Finn2017ModelAgnosticMF, snell-prototypical, ravi2018amortized, Nichol2018OnFM}. \citet{mamlenllm}  adapted Model-Agnostic Meta-Learning \citep{Finn2017ModelAgnosticMF} to LLMs. However, this adaptation is more expensive in computation and memory than our method, requiring second-order gradient updates and saving a model for each task. More recently, \citet{lift2025} proposed a hierarchical Bayesian approach to LoRA meta-learning, but its parameters also increase linearly with number of tasks. As a result, we are able to evaluate performance on larger models than the ones tried in these two papers.

In a different approach, \citet{min-etal-2022-metaicl} and \citet{chen-etal-2022-meta} explored meta-learning for LLMs using in-context learning (ICL). These works show that it is possible to fine-tune LLMs on in-context examples and achieve generalization. However, ICL has shown weaknesses in utilizing knowledge that are both novel and complex \citep{dou2026clbenchbenchmarkcontextlearning}. We show that in-weights meta-learning can utilize ICL to achieve a better performance.

\paragraph{Uncertainty representation for LLMs.} 
Approaches to capturing uncertainty for LLMs can rely on the intrinsic representation of uncertainty in the model or focus on capturing extrinsic uncertainty about model parameters. Intrinsic approaches produce better uncertainty calibration via prompt engineering and sampling \citep{gruver-time-series} or learning an external model \citep{pmlr-v235-shen24c}. Extrinsic approaches include using fine-tuning methods to incorporate uncertainty, such as training LoRA with ensembles \citep{Balabanov2024UncertaintyQI}, Laplace approximation \citep{Yang2023BayesianLA}, and variational inference \citep{wang2024blob}. Our work takes the extrinsic approach but differs from existing approaches by using meta-learning to achieve generalization across datasets.

\section{Background}

\subsection{Low-Rank Adaptation (LoRA)}

LoRA \citep{Hu2021LoRALA} fine-tunes LLM weights on a low-rank space to improve efficiency compared with regular fine-tuning. Let $\mathbf{W}_0$ of size $d_\text{out}$-by-$d_\text{in}$ denote a weight matrix from a pretrained LLM. Let $\mathbf{x}$ denote the input to $\mathbf{W}_0$, and $\mathbf{z}$ denote the output of $\mathbf{W}_0$, LoRA fine-tunes the pretrained weight $\mathbf{W}_0$ by adding a low-rank matrix that is the product of two trainable matrices,
\begin{align*}
    \mathbf{h} = (\mathbf{W}_0 + \Delta \mathbf{W}_0)\mathbf{x} =(\mathbf{W}_0 + \mathbf{B}\mathbf{A})\mathbf{x}.
\end{align*}
The trainable matrices $\mathbf{B}$ and $\mathbf{A}$ are known as \textit{LoRA adapters}. The sizes of $\mathbf{B}$ and $\mathbf{A}$ are $d_\text{out}$-by-$d_\text{rank}$ and $d_\text{rank}$-by-$d_\text{in}$, respectively, with $d_\text{rank}$ being significantly smaller than the original dimensions. Therefore, the number of parameters to be updated are $(d_\text{out} + d_\text{in})d_\text{rank}$, significantly fewer than the original $d_\text{out}d_\text{in}$.

\subsection{Approaches to Meta-Learning}

Meta-learning aims to find a set of initial model parameters that can be rapidly adapted to new, unseen tasks with a few gradient steps \citep{schmidhuber1987evolutionary,bengio1991learning,caruana1998multitask}. The strategy for doing so is to generalize from the shared statistical structure across tasks: by extracting this structure, a model can ``learn to learn.'' A common setting for meta-learning is few-shot learning, where each task has few examples and there are many such tasks. Personalizing large language models through modification of their weights, rather than their prompts or the context they condition on, is a natural setting for using this approach, where each user might only have a limited amount of data available but a group of users may have similar interests.

\paragraph{MAML.} A popular approach to meta-learning is the Model-Agnostic Meta-Learning \citep[MAML;][]{Finn2017ModelAgnosticMF} algorithm. This algorithm runs two training loops: an inner loop for task-specific adaptation and an outer loop for meta-optimization. Let $D_i$ denote a batch of data from task $i$, $p_\theta$ denote the prediction model parameterized by $\theta$, and $\alpha,\beta$ denote gradient descent step sizes. The goal of learning is to obtain a set of parameters $\theta_i$ for each task and a global set of parameters $\theta$ that are used to initialize learning for all tasks. In a given epoch, for each task $i$, MAML conducts an inner loop gradient update,
\[
    \theta_i = \theta -\alpha \nabla_{\theta} \mathcal{L}_{D_i}(p_{\theta})
\]
where $\mathcal{L}$ denotes the loss function, e.g. the cross-entropy loss. After executing a set of inner loops the outer loop update is executed,
\begin{align*}
    \theta \leftarrow \theta -\beta \nabla_{\theta} \sum_{i} \mathcal{L}_{D_i}(p_{\theta_i^{'}}).
\end{align*}
With the outer loop update, MAML is trained to find a more generalizable set of parameters $\theta$ that is ``close'' to the optimal parameters for many tasks. The downside of MAML is the computational and memory requirements  that can be seen in these updates. A copy of the model parameters $\theta_i$ must be cached for each task $i$. Additionally, the outer loop updates feature a gradient over the gradient of $\theta$, thus requiring a second-order gradient update.

\paragraph{Reptile.}

Reptile simplifies and approximates the approach to meta-learning adopted in MAML. For each task $i$, it updates the current parameters $\theta$ $k$ times, each time as a regular stochastic gradient descent update,
\begin{align*}
    \theta_i \leftarrow \theta -\alpha \nabla_{\theta}\mathcal{L}_{D_i}(p_{\theta})
\end{align*}
After these $k$ updates, a meta-update is used to improve the global parameters $\theta$,
\begin{align*}
    \theta \leftarrow \theta + \epsilon (\theta_i - \theta),
\end{align*}
with $0<\epsilon < 1$. This can be interpreted as a gradient descent procedure where $\theta - \theta_i$ is taken as the gradient. An advantage of Reptile is efficiency: scaling with respect to the number of tasks, and not having a second-order gradient update.

\subsection{Amortized Bayesian Meta-Learning}

Amortized Bayesian Meta-Learning \citep[ABML;][]{ravi2018amortized} improves upon MAML-based meta-learning frameworks by representing uncertainty with a Bayesian approach. It also amortizes inference over the parameters so that memory no longer increases linearly with the number of tasks. 

Let $\theta$ denote global parameters such that a few steps of gradient descent will produce local parameters $\phi_i$ on task $i$ with dataset $D_i$.
ABML treats $\theta$ as random variables, and minimizes a negative evidence lower bound using variational inference,
\begin{equation}
\underset{\theta}{\text{argmin}}\sum_{i=1}^M \left[-\mathbb{E}_{q_{\theta}(\phi_i|D_i)}[\log p(D_i|\phi_i)] + \text{KL}\big(q_{\theta}(\phi_i|D_i) \,\|\, p(\phi_i|\theta) \big) \right] + \text{KL}(q(\theta) \,\|\, p(\theta)). \label{eq:obj}
\end{equation}
The variational distribution $q_{\theta}(\phi_i|D_i)$ is represented by the Gaussian distribution $N(\mu_{\phi}, \sigma_{\phi}^2)$ with $\mu_{\phi}, \sigma_{\phi}$ as trainable parameters. 





\section{Method}

Meta-learning enables models to develop more generalizable learning strategies. Yet, due to its computational overhead, it is underexplored on larger LLMs with billions of parameters. Our method, Amortized Bayesian Meta-Learning for LoRA (ABMLL), extends ABML, making it possible to apply to LLMs. This approach combines the advantages of meta-learning for adapting to new tasks with Bayesian probabilistic modeling for instantiating this idea and for representing uncertainty. 


ABMLL uses the the objective of Eq. \ref{eq:obj} from ABML.
In our setting, $\theta$ and $\phi_i$ are the global and task-specific model parameters produced as the output of LoRA adapters. At a high level, the generative process is
\begin{align*}
    \theta &\sim p(\theta), \;\; \phi_i \sim p(\phi_i|\theta), \;\; D_i \sim \text{LLM}(\phi_i),
\end{align*}
where $i$ indexes over tasks and $\text{LLM}(\phi_i)$ denotes the LLM considered as a probabilistic model that takes $\phi_i$ as its inputs and outputs token sequences with joint probabilities defined by the LLM's autoregressive predictive distribution. By positing that task-specific variables $\phi_i$ are generated from global variables $\theta$, the model is encouraged to learn a generalizable space of parameters with fast adaptions to different tasks. More detailed algorithm is given in the Appendix \ref{sec:appendix-alg}, Algorithm \ref{alg:abmll}.
For any LLM layer with pretrained weights $\mathbf{W_0}$, the quantities for our extension to ABML are:
\vspace{-0mm}    
\[
\begin{array}{cc}
\begin{aligned}[t]
\mu_\theta &= \mathbf{B}_{\mu_{\theta}}\mathbf{A}_{\mu_{\theta}}, \\
\log\sigma_\theta^2 &= \mathbf{B}_{\sigma_{\theta}}\mathbf{A}_{\sigma_{\theta}} + c\mathbf{I}, \\
\mu_\phi &= \mathbf{B}_{\mu_{\phi}}\mathbf{A}_{\mu_{\phi}}, \\
\log \sigma_\phi^2 &= \mathbf{B}_{\sigma_{\phi}}\mathbf{A}_{\sigma_{\phi}} + c\mathbf{I}
\end{aligned}
&
\begin{aligned}[t]
p(\phi_i \mid \theta) &= \mathcal{N}(\phi_i ; \mu_{\theta}+\mathbf{W_0}, \sigma_{\theta}^2), \\
q_{\theta}(\phi_i \mid D_i) &= \mathcal{N}(\phi_i; \mu_{\phi}+\mathbf{W_0}, \sigma_{\phi}^2), \\
p(\theta) &= p(\mu_{\theta}, \sigma_{\theta}) = \mathcal{N}(\mu_{\theta} ; 0, \mathbf{I}) \cdot \text{Gamma}\left(\frac{1}{\sigma_{\theta}^2}; a_0, b_0\right), \\
\text{KL}(q(\theta) &\,\|\, p(\theta)) = -\log p(\theta)
\end{aligned}
\end{array}
\]

Lastly, $p(D_i|\phi_i)$ is defined as the joint probability assigned to $D_i$ where the LLM takes $\phi_i$ as its weights. The trainable parameters are the LoRA adapters $\mathbf{A}$ and $\mathbf{B}$. However, we introduce four pairs of these adapters to compute both the mean and variance of the LoRA outputs on local and global model weights. $\mathbf{I}$ is identity matrix, and $c$ is a hyperparameter constant dependent on the spread of pretrained LLM weights. $a_0$ and $b_0$ are hyperparameters, and the simplification of the KL term as $-\log p(\theta)$ follows \citet{ravi2018amortized}.

\paragraph{Balancing the reconstruction error.} LLMs are often overparameterized. As a result, probabilistic quantities on the space of weights, $\text{KL}\big(q_{\theta}(\phi_i|D_i) \big| \big| p(\phi_i|\theta) \big)$ and $\text{KL}(q(\theta) ||p(\theta))$, can overwhelm quantities on the data space, $\log p(D_i|\phi_i)$. $\beta$-VAE \citep{Higgins2016betaVAELB} and Bayesian neural network approaches \citep{trinh2022tackling} introduce hyperparameters to temper the likelihood versus regularization terms. Inspired by this idea, we introduce hyperparameter $\beta$, resulting in the following objective,
\vspace{-0mm}
\begin{equation}
\underset{\theta}{\text{argmin}} \sum_{i=1}^M \left[  -\mathbb{E}_{q_{\theta}(\phi_i|D_i)}[\log p(D_i|\phi_i)] ] + \beta\text{KL}\left(q_{\theta}(\phi_i|D_i) \, \| \, p(\phi_i|\theta) \right) \right] + \beta \text{KL}(q(\theta) \,\|\,p(\theta)). \label{eq:main}
\end{equation}
This provides a flexible way to control how close the global parameters $\theta$ are to the prior $p(\theta)$, and how close the task-specific parameters $\phi_i$ are to $\theta$.

\section{Empirical Evaluations}

\subsection{Performance on Few-Shot Learning Datasets}

We examine few-shot learning performance by meta-learning methods on natural corpora.

\subsubsection{Experimental Setup}
\label{sec:experimental-setup}

\paragraph{Model and datasets.} We fine-tune \textsc{Llama3-8B} and \textsc{Qwen2-7B} on CrossFit and UnifiedQA \citep{ye-etal-2021-crossfit}, two text datasets commonly used to train meta-learning models. 

We train and evaluate in three settings. The first setting is cls-45, where models are trained on classification tasks, and evaluated on other distinct classification tasks. The second setting is cls-23, where models are evaluated on the same classification tasks as in cls-45, but are trained on a mix of classification and other tasks including question-answering and natural language inference. Finally, following \citet{min-etal-2022-metaicl}, we test stronger generalizations on more narrowly defined tasks, where models train on other tasks, but evaluate on only natural language inference (NLI), paraphrasing (Para), and knowledge-based multiple choice questions-answers (MCQA), respectively.

Because one aim of our paper is to study uncertainty quantification, we focus on multiple choice datasets. In the case of cls-45, this results in a subset of CrossFit and UnifiedQA with 34 training tasks, 15 evaluation tasks, and 68K training datapoints in total. For more details on datasets, see Section \ref{sec:appendix-data} in the Appendix. 


\paragraph{Metrics.} We use accuracy to evaluate general performance and expected calibration error (ECE) to evaluate uncertainty estimation. 

\paragraph{Baselines.} We use five baseline methods that can viably scale to our context. \textit{Pretrained} is the off-the-shelf LLM. \textit{Regular LoRA} is the default LoRA method trained on the whole randomly shuffled training dataset. \textit{Structured LoRA} also uses the default LoRA, but the training dataset follows the same ``structure'' as our method: it is iteratively trained 5 gradient steps on one task at a time, dropping the global variable. Thus, it tests the effect of our generative model on performance. \textit{Reptile} \citep{Nichol2018OnFM} implements the Reptile meta-learning algorithm. 

\paragraph{Implementation details.} ABMLL takes one sample from the reparameterization step during inference. During validation on the unseen dataset, all models perform 5-shot learning: training 5 gradient steps on 5 batches from this dataset and evaluating on the rest. Appendix \ref{sec:appendix-exp} includes more details.

\begin{figure*}[t]
    \centering
    \subfigure{
        \centering
        \includegraphics[width=0.46\textwidth]{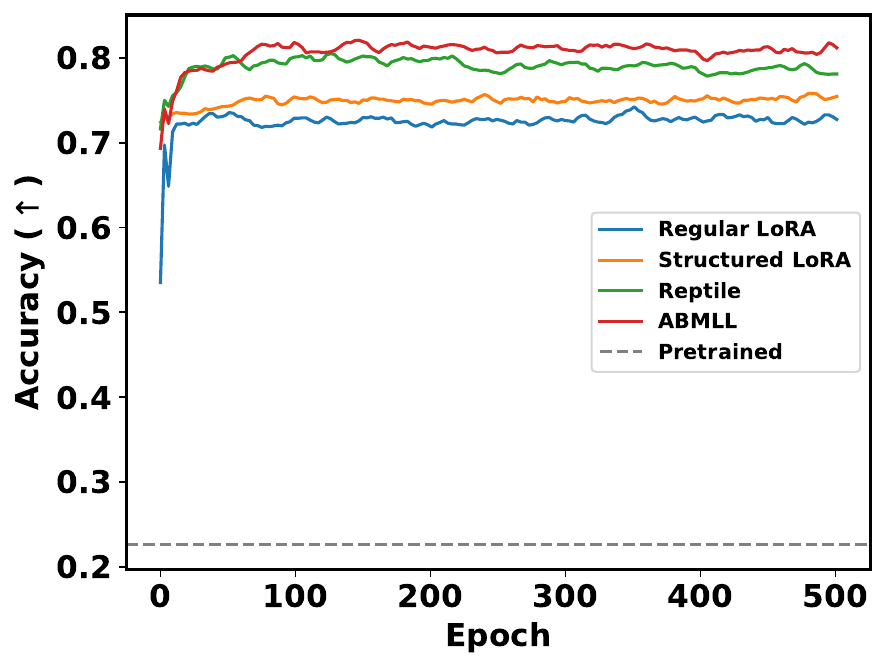}
        } 
    \subfigure{
        \centering
        \includegraphics[width=0.47\textwidth]{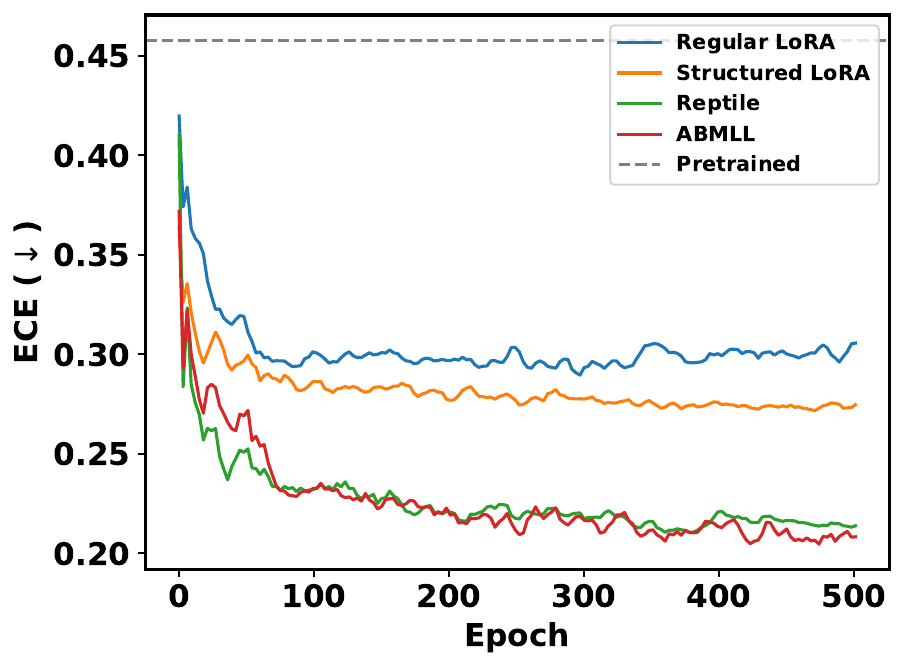}
        }
    \caption{cls-45 validation accuracy and ECE over epochs across our method (ABMLL) and four benchmarks. Values are computed as sliding-window moving average over the three most recent epochs. ABMLL achieves consistent performance on both metrics.}
    \label{fig:across-epochs}
\end{figure*}

\subsubsection{Experimental Results - CrossFit and UnifiedQA}

Figure \ref{fig:across-epochs} shows accuracy and ECE over epochs on cls-45. We observe that the meta-learning methods (ABMLL and Reptile) achieve evidently higher accuracy than methods not based on meta-learning. Among the two meta-learning methods, the performance of ABMLL is slightly but consistently better than Reptile. 

\paragraph{Meta-learning outperforms its counterparts without meta-learning.} Table \ref{tab:acc-ece-nonicl} reports test scores for each model from three random seeds, where the test results are taken from each model's best validation accuracy epoch. For methods without in-context examples, ABMLL performs best on both metrics on cls-45 and cls-23, suggesting that ABMLL trains a general learner able to adapt to a variety of problems. The performance of ABMLL is tied with Reptile on tasks with more distinct train-evaluation differences, possibly because the tested ability is more difficult to acquire through training tasks, bringing the performance of these two meta-learning models closer together. Meanwhile, the two meta-learning methods perform significantly better than the rest, suggesting that meta-learning is an important ingredient for fine-tuning LLMs with stronger generalization.

\paragraph{Memory consumption.}

An advantage of ABMLL over many meta-learning methods is scalability. Although ABMLL introduces four pairs of LoRA adapters, pretrained weights from both ABMLL and LoRA still need to be computed during a forward pass, despite having no gradients attached to them. ABMLL increases peak memory by only $7.6\%$ over regular LoRA on cls-45.

\begin{table}[t]
\centering
\caption{Test accuracy and ECE across three random seeds, with standard error, for two models (\textsc{Llama3} and \textsc{Qwen2}). Best performance is shown in bold.}
\resizebox{0.95\columnwidth}{!}{
\begin{tabular}{llccccc}
 \toprule
 Model & Method & cls-45 Acc $\uparrow$ & cls-23 Acc $\uparrow$ & NLI Acc $\uparrow$ & Para Acc $\uparrow$ & MCQA Acc $\uparrow$ \\
 \midrule

 \multirow{5}{*}{\textsc{Llama3}}
 & Pretrained
 & $26.1\%$ \scriptsize{$\pm 0.1\%$}
 & $26.0\%$ \scriptsize{$\pm 0.1\%$}
 & $57.6\%$ \scriptsize{$\pm 0.0\%$}
 & $57.0\%$ \scriptsize{$\pm 0.0\%$}
 & $71.8\%$ \scriptsize{$\pm 0.1\%$} \\
 & Regular LoRA
 & $71.6\%$ \scriptsize{$\pm 0.3\%$}
 & $71.4\%$ \scriptsize{$\pm 0.5\%$}
 & $78.5\%$ \scriptsize{$\pm 0.0\%$}
 & $59.9\%$ \scriptsize{$\pm 0.4\%$}
 & $76.1\%$ \scriptsize{$\pm 0.1\%$} \\
 & Struct. LoRA
 & $74.5\%$ \scriptsize{$\pm 0.1\%$}
 & $71.4\%$ \scriptsize{$\pm 0.0\%$}
 & $75.5\%$ \scriptsize{$\pm 0.1\%$}
 & $55.1\%$ \scriptsize{$\pm 0.1\%$}
 & $76.0\%$ \scriptsize{$\pm 0.0\%$} \\
 & Reptile
 & $73.0\%$ \scriptsize{$\pm 0.3\%$}
 & $72.7\%$ \scriptsize{$\pm 0.2\%$}
 & \underline{\bm{$83.3\%$}} \scriptsize{$\pm 0.3\%$}
 & \underline{\bm{$61.8\%$}} \scriptsize{$\pm 0.2\%$}
 & \underline{\bm{$77.7\%$}} \scriptsize{$\pm 0.1\%$} \\
 & ABMLL (ours)
 & \underline{\bm{$75.2\%$}} \scriptsize{$\pm 0.0\%$}
 & \underline{\bm{$73.3\%$}} \scriptsize{$\pm 0.1\%$}
 & $82.2\%$ \scriptsize{$\pm 0.1\%$}
 & \underline{\bm{$61.6\%$}} \scriptsize{$\pm 1.9\%$}
 & $77.4\%$ \scriptsize{$\pm 0.1\%$} \\

 \midrule
 \multirow{5}{*}{\textsc{Qwen2}}
 & Pretrained
 & $60.5\%$ \scriptsize{$\pm 0.1\%$}
 & $60.5\%$ \scriptsize{$\pm 0.1\%$}
 & $69.3\%$ \scriptsize{$\pm 0.1\%$}
 & $57.1\%$ \scriptsize{$\pm 0.1\%$}
 & $77.5\%$ \scriptsize{$\pm 0.2\%$} \\
 & Regular LoRA
 & $66.3\%$ \scriptsize{$\pm 0.3\%$}
 & $66.0\%$ \scriptsize{$\pm 0.2\%$}
 & $79.7\%$ \scriptsize{$\pm 0.3\%$}
 & $63.3\%$ \scriptsize{$\pm 0.9\%$}
 & $77.5\%$ \scriptsize{$\pm 0.2\%$} \\
 & Struct. LoRA
 & $69.6\%$ \scriptsize{$\pm 0.2\%$}
 & $67.4\%$ \scriptsize{$\pm 0.2\%$}
 & $79.1\%$ \scriptsize{$\pm 0.2\%$}
 & $62.2\%$ \scriptsize{$\pm 0.4\%$}
 & $77.8\%$ \scriptsize{$\pm 0.2\%$} \\
 & Reptile
 & $72.9\%$ \scriptsize{$\pm 0.1\%$}
 & $70.0\%$ \scriptsize{$\pm 0.4\%$}
 & $82.6\%$ \scriptsize{$\pm 0.5\%$}
 & $63.4\%$ \scriptsize{$\pm 0.2\%$}
 & \underline{\bm{$79.0\%$}} \scriptsize{$\pm 0.1\%$} \\
 & ABMLL (ours)
 & \underline{\bm{$74.1\%$}} \scriptsize{$\pm 0.7\%$}
 & \underline{\bm{$72.4\%$}} \scriptsize{$\pm 0.3\%$}
 & \underline{\bm{$83.6\%$}} \scriptsize{$\pm 0.0\%$}
 & \underline{\bm{$66.1\%$}} \scriptsize{$\pm 0.9\%$}
 & $77.2\%$ \scriptsize{$\pm 0.7\%$} \\

 \toprule
 Model & Method & cls-45 ECE $\downarrow$ & cls-23 ECE $\downarrow$ & NLI ECE $\downarrow$ & Para ECE $\downarrow$ & MCQA ECE $\downarrow$ \\
 \midrule

 \multirow{5}{*}{\textsc{Llama3}}
 & Pretrained
 & $0.458$ \scriptsize{$\pm 0.000$}
 & $0.458$ \scriptsize{$\pm 0.000$}
 & $0.419$ \scriptsize{$\pm 0.000$}
 & $0.430$ \scriptsize{$\pm 0.000$}
 & $0.276$ \scriptsize{$\pm 0.000$} \\
 & Regular LoRA
 & $0.318$ \scriptsize{$\pm 0.001$}
 & $0.328$ \scriptsize{$\pm 0.006$}
 & $0.310$ \scriptsize{$\pm 0.003$}
 & $0.433$ \scriptsize{$\pm 0.002$}
 & $0.286$ \scriptsize{$\pm 0.003$} \\
 & Struct. LoRA
 & $0.288$ \scriptsize{$\pm 0.001$}
 & $0.305$ \scriptsize{$\pm 0.001$}
 & $0.302$ \scriptsize{$\pm 0.001$}
 & $0.477$ \scriptsize{$\pm 0.001$}
 & $0.290$ \scriptsize{$\pm 0.000$} \\
 & Reptile
 & $0.278$ \scriptsize{$\pm 0.000$}
 & $0.284$ \scriptsize{$\pm 0.001$}
 & \underline{\bm{$0.242$}} \scriptsize{$\pm 0.004$}
 & \underline{\bm{$0.404$}} \scriptsize{$\pm 0.003$}
 & $0.277$ \scriptsize{$\pm 0.001$} \\
 & ABMLL (ours)
 & \underline{\bm{$0.262$}} \scriptsize{$\pm 0.001$}
 & \underline{\bm{$0.273$}} \scriptsize{$\pm 0.005$}
 & \underline{\bm{$0.237$}} \scriptsize{$\pm 0.020$}
 & \underline{\bm{$0.413$}} \scriptsize{$\pm 0.007$}
 & \underline{\bm{$0.275$}} \scriptsize{$\pm 0.000$} \\

 \midrule
 \multirow{5}{*}{\textsc{Qwen2}}
 & Pretrained
 & $0.397$ \scriptsize{$\pm 0.000$}
 & $0.397$ \scriptsize{$\pm 0.000$}
 & $0.298$ \scriptsize{$\pm 0.000$}
 & $0.428$ \scriptsize{$\pm 0.000$}
 & \underline{\bm{$0.223$}} \scriptsize{$\pm 0.000$} \\
 & Regular LoRA
 & $0.322$ \scriptsize{$\pm 0.001$}
 & $0.330$ \scriptsize{$\pm 0.001$}
 & $0.269$ \scriptsize{$\pm 0.001$}
 & $0.423$ \scriptsize{$\pm 0.001$}
 & $0.241$ \scriptsize{$\pm 0.002$} \\
 & Struct. LoRA
 & $0.307$ \scriptsize{$\pm 0.000$}
 & $0.321$ \scriptsize{$\pm 0.001$}
 & $0.270$ \scriptsize{$\pm 0.002$}
 & $0.401$ \scriptsize{$\pm 0.002$}
 & $0.239$ \scriptsize{$\pm 0.002$} \\
 & Reptile
 & $0.279$ \scriptsize{$\pm 0.001$}
 & $0.293$ \scriptsize{$\pm 0.002$}
 & $0.231$ \scriptsize{$\pm 0.003$}
 & $0.373$ \scriptsize{$\pm 0.001$}
 & $0.233$ \scriptsize{$\pm 0.002$} \\
 & ABMLL (ours)
 & \underline{\bm{$0.273$}} \scriptsize{$\pm 0.009$}
 & \underline{\bm{$0.285$}} \scriptsize{$\pm 0.006$}
 & \underline{\bm{$0.208$}} \scriptsize{$\pm 0.008$}
 & \underline{\bm{$0.364$}} \scriptsize{$\pm 0.004$}
 & $0.249$ \scriptsize{$\pm 0.004$} \\

 \bottomrule
\end{tabular}}
\label{tab:acc-ece-nonicl}
\end{table}

\begin{table}[t]
\centering
\caption{Test accuracy and ECE across three random seeds, with standard error, for two models (\textsc{Llama3} and \textsc{Qwen2}) in the ICL setting. Best performances is shown in bold.}
\resizebox{1\columnwidth}{!}{
\begin{tabular}{llccccc}
 \toprule
 Model & Method & cls-45 Acc $\uparrow$ & cls-23 Acc $\uparrow$ & NLI Acc $\uparrow$ & Para Acc $\uparrow$ & MCQA Acc $\uparrow$ \\
 \midrule

 \multirow{3}{*}{\textsc{Llama3}}
 & MetaICL 
 & $71.0\%$ \scriptsize{$\pm 0.1\%$} 
 & $70.8\%$ \scriptsize{$\pm 0.2\%$} 
 & $83.3\%$ \scriptsize{$\pm 0.3\%$} 
 & $61.6\%$ \scriptsize{$\pm 0.2\%$} 
 & \underline{\bm{$80.1\%$}} \scriptsize{$\pm 0.2\%$} \\
 & Reptile-MetaICL 
 & $74.2\%$ \scriptsize{$\pm 0.8\%$} 
 & $73.5\%$ \scriptsize{$\pm 0.2\%$} 
 & $83.6\%$ \scriptsize{$\pm 0.2\%$} 
 & $64.0\%$ \scriptsize{$\pm 0.8\%$} 
 & \underline{\bm{$80.2\%$}} \scriptsize{$\pm 0.2\%$} \\
 & ABMLL-MetaICL (ours)
 & \underline{\bm{$77.4\%$}} \scriptsize{$\pm 0.5\%$} 
 & \underline{\bm{$75.4\%$}} \scriptsize{$\pm 0.6\%$} 
 & \underline{\bm{$85.2\%$}} \scriptsize{$\pm 0.1\%$} 
 & \underline{\bm{$68.1\%$}} \scriptsize{$\pm 0.3\%$} 
 & \underline{\bm{$80.2\%$}} \scriptsize{$\pm 0.3\%$} \\

 \midrule
 \multirow{3}{*}{\textsc{Qwen2}}
 & MetaICL 
 & $72.2\%$ \scriptsize{$\pm 0.5\%$} 
 & $70.2\%$ \scriptsize{$\pm 0.3\%$} 
 & \underline{\bm{$83.8\%$}} \scriptsize{$\pm 0.2\%$} 
 & $76.2\%$ \scriptsize{$\pm 0.2\%$} 
 & $81.3\%$ \scriptsize{$\pm 0.1\%$} \\
 & Reptile-MetaICL 
 & $74.9\%$ \scriptsize{$\pm 0.9\%$} 
 & \underline{\bm{$74.2\%$}} \scriptsize{$\pm 0.6\%$} 
 & $83.0\%$ \scriptsize{$\pm 0.1\%$} 
 & $77.8\%$ \scriptsize{$\pm 0.8\%$} 
 & \underline{\bm{$82.2\%$}} \scriptsize{$\pm 0.3\%$} \\
 & ABMLL-MetaICL (ours)
 & \underline{\bm{$77.2\%$}} \scriptsize{$\pm 0.2\%$} 
 & \underline{\bm{$75.3\%$}} \scriptsize{$\pm 0.7\%$} 
 & $82.6\%$ \scriptsize{$\pm 0.3\%$} 
 & \underline{\bm{$80.6\%$}} \scriptsize{$\pm 0.3\%$} 
 & \underline{\bm{$82.1\%$}} \scriptsize{$\pm 0.1\%$} \\

 \toprule
 Model & Method & cls-45 ECE $\downarrow$ & cls-23 ECE $\downarrow$ & NLI ECE $\downarrow$ & Para ECE $\downarrow$ & MCQA ECE $\downarrow$ \\
 \midrule

 \multirow{3}{*}{\textsc{Llama3}}
 & MetaICL 
 & $0.301$ \scriptsize{$\pm 0.002$} 
 & $0.307$ \scriptsize{$\pm 0.001$} 
 & $0.246$ \scriptsize{$\pm 0.008$} 
 & $0.411$ \scriptsize{$\pm 0.002$} 
 & $0.251$ \scriptsize{$\pm 0.001$} \\
 & Reptile-MetaICL 
 & $0.279$ \scriptsize{$\pm 0.008$} 
 & \underline{\bm{$0.263$}} \scriptsize{$\pm 0.007$} 
 & $0.281$ \scriptsize{$\pm 0.014$} 
 & \underline{\bm{$0.392$}} \scriptsize{$\pm 0.006$} 
 & \underline{\bm{$0.246$}} \scriptsize{$\pm 0.002$} \\
 & ABMLL-MetaICL (ours)
 & \underline{\bm{$0.241$}} \scriptsize{$\pm 0.000$} 
 & \underline{\bm{$0.271$}} \scriptsize{$\pm 0.003$} 
 & \underline{\bm{$0.207$}} \scriptsize{$\pm 0.006$} 
 & \underline{\bm{$0.402$}} \scriptsize{$\pm 0.017$} 
 & \underline{\bm{$0.244$}} \scriptsize{$\pm 0.006$} \\
 \midrule
 \multirow{3}{*}{\textsc{Qwen2}}
 & MetaICL 
 & $0.292$ \scriptsize{$\pm 0.004$} 
 & $0.295$ \scriptsize{$\pm 0.001$} 
 & $0.246$ \scriptsize{$\pm 0.002$} 
 & $0.315$ \scriptsize{$\pm 0.002$} 
 & \underline{\bm{$0.213$}} \scriptsize{$\pm 0.001$} \\
 & Reptile-MetaICL 
 & $0.278$ \scriptsize{$\pm 0.008$} 
 & \underline{\bm{$0.287$}} \scriptsize{$\pm 0.003$} 
 & $0.252$ \scriptsize{$\pm 0.001$} 
 & $0.314$ \scriptsize{$\pm 0.002$} 
 & \underline{\bm{$0.212$}} \scriptsize{$\pm 0.003$} \\
 & ABMLL-MetaICL (ours)
 & \underline{\bm{$0.249$}} \scriptsize{$\pm 0.005$} 
 & \underline{\bm{$0.284$}} \scriptsize{$\pm 0.002$} 
 & \underline{\bm{$0.227$}} \scriptsize{$\pm 0.015$} 
 & \underline{\bm{$0.288$}} \scriptsize{$\pm 0.012$} 
 & $0.217$ \scriptsize{$\pm 0.001$} \\
 \bottomrule
\end{tabular}}
\label{tab:acc-ece-icl}
\end{table}

\begin{table}[t]
\centering
\caption{Extension of Table \ref{tab:acc-ece-icl} to LegalBench and ChemBench datasets, two domain applications that use ICL by default.}
\resizebox{1\columnwidth}{!}{
\begin{tabular}{llcccc}
 \toprule
 Model & Method & LegalBench Acc $\uparrow$ & ChemBench Acc $\uparrow$ & LegalBench ECE $\downarrow$ & ChemBench ECE $\downarrow$ \\
 \midrule
 \multirow{3}{*}{\textsc{Llama3}}
 & MetaICL 
 & $77.1\%$ \scriptsize{$\pm 0.1\%$} 
 & \underline{\bm{$55.0\%$}} \scriptsize{$\pm 0.4\%$}
 & $0.265$ \scriptsize{$\pm 0.002$} 
 & $0.403$ \scriptsize{$\pm 0.001$} \\
 & Reptile-MetaICL 
 & $79.1\%$ \scriptsize{$\pm 0.2\%$} 
 & $54.6\%$ \scriptsize{$\pm 0.4\%$}
 & \underline{\bm{$0.261$}} \scriptsize{$\pm 0.004$} 
 & \underline{\bm{$0.393$}} \scriptsize{$\pm 0.003$} \\
 & ABMLL-MetaICL (ours)
 & \underline{\bm{$79.5\%$}} \scriptsize{$\pm 0.1\%$} 
 & \underline{\bm{$55.5\%$}} \scriptsize{$\pm 0.1\%$}
 & $0.285$ \scriptsize{$\pm 0.006$} 
 & \underline{\bm{$0.394$}} \scriptsize{$\pm 0.004$} \\
 \midrule
 \multirow{3}{*}{\textsc{Qwen2}}
 & MetaICL 
 & $75.7\%$ \scriptsize{$\pm 0.1\%$} 
 & $53.4\%$ \scriptsize{$\pm 0.0\%$}
 & $0.291$ \scriptsize{$\pm 0.004$} 
 & $0.417$ \scriptsize{$\pm 0.001$} \\
 & Reptile-MetaICL 
 & \underline{\bm{$78.8\%$}} \scriptsize{$\pm 0.1\%$} 
 & \underline{\bm{$58.1\%$}} \scriptsize{$\pm 0.1\%$}
 & $0.252$ \scriptsize{$\pm 0.003$} 
 & \underline{\bm{$0.394$}} \scriptsize{$\pm 0.001$} \\
 & ABMLL-MetaICL (ours)
 & $78.3\%$ \scriptsize{$\pm 0.2\%$} 
 & \underline{\bm{$58.4\%$}} \scriptsize{$\pm 0.9\%$}
 & \underline{\bm{$0.246$}} \scriptsize{$\pm 0.002$} 
 & \underline{\bm{$0.397$}} \scriptsize{$\pm 0.003$} \\
 \bottomrule
\end{tabular}}
\label{tab:domain-acc-ece}
\end{table}

\subsection{Combining Meta-Learning with ICL}

Meta-learning is a complementary approach to ICL, and being able to use meta-learning at scale makes it possible to combined these techniques. We compare
using MetaICL \citep{min-etal-2022-metaicl} and augmenting Reptile and ABMLL with in-context learning.

This section extends the evaluation datasets to two more domain applications with in-context learning - legal and chemistry domains \citep{guha2023legalbench, mirza2024chembench}. For these evaluations, the training data is the same as the previous cls45 training split, a split that is generic for CrossFit and UnifiedQA. We hypothesize that even if the training data were distant from legal and chemistry domains, the weight adaptability encouraged by meta-learning should speed up few-shot learning for these domains.

\subsubsection{Experimental Results - CrossFit, UnifiedQA, LegalBench, ChemBench}

Meta-learning outperforms default LoRA to an even greater extent when combined with in-context examples, as demonstrated by the performance of \textit{ABMLL-MetaICL} and \textit{Reptile-MetaICL} (Table \ref{tab:acc-ece-icl}). \textit{ABMLL} shows the strongest synergy with ICL, outperforming regular LoRA by over $10\%$ in accuracy on several datasets, and significantly outperforming regular MetaICL as well. The two meta-learning methods also generalize better to LegalBench and ChemBench (Table \ref{tab:domain-acc-ece}).  These results suggests that meta-learning provides complementary gains to ICL, with the potential to improve performance in settings where ICL is  used.

\subsection{Model Pruning}

Model pruning improves efficiency and probes robustness by removing redundant parameters, potentially eliminating spurious correlations \citep{Ashkboos2024SliceGPTCL, Sun2023ASA}. Bayesian neural networks are known to be resource-efficient \citep{blundell_2015}. Inspired by this idea, we test the performance of ABMLL and benchmarks by setting a percentage of neurons in each layer embedding to zero, sorted by magnitudes of these neurons.

Table \ref{tab:prune} shows that ABMLL is significantly more robust against this form of pruning than the other methods. This result suggests that ABMLL is robust and reliable, learning generalizable features that are not as tied to specific parameters. Although weight decay can improve pruning performance relative to no regularization, it hurts the unpruned setting and still underperforms ABMLL under pruning.

\begin{table}[t]
\centering
\caption{Pruning results across methods on two datasets. In each column except the first, a certain percentage of neurons in each layer embedding is set to zero. ABMLL is significantly more robust against pruning than the other methods. For more datasets, see Appendix \ref{sec:appendix-exp}.} 
\resizebox{0.82\textwidth}{!}{
\subtable[NLI.]{
\begin{tabular}{llllll} 
 \toprule
 Method & $0\%$ Pruned & $1\%$ Pruned & $10\%$ Pruned & $20\%$ Pruned & $30\%$ Pruned\\
 \midrule
 Pretrained 
 & $57.6\%$ \scriptsize{$\pm 0.0\%$}
 & $47.9\%$ \scriptsize{$\pm 0.0\%$}
 & $48.2\%$ \scriptsize{$\pm 0.0\%$}
 & $48.2\%$ \scriptsize{$\pm 0.0\%$}
 & $43.6\%$ \scriptsize{$\pm 0.2\%$}\\
 Regular LoRA 
 & $78.5\%$ \scriptsize{$\pm 0.0\%$}
 & $69.2\%$ \scriptsize{$\pm 0.4\%$}
 & $68.4\%$ \scriptsize{$\pm 0.3\%$}
 & $66.7\%$ \scriptsize{$\pm 0.5\%$}
 & $60.9\%$ \scriptsize{$\pm 0.6\%$}\\
 Struct. LoRA 
 & $75.5\%$ \scriptsize{$\pm 0.1\%$}
 & $74.0\%$ \scriptsize{$\pm 0.2\%$}
 & $73.9\%$ \scriptsize{$\pm 0.2\%$}
 & $73.4\%$ \scriptsize{$\pm 0.2\%$}
 & $64.6\%$ \scriptsize{$\pm 0.2\%$}\\
 Reptile 
 & \bm{$83.3\%$} \scriptsize{$\pm 0.3\%$}
 & $78.3\%$ \scriptsize{$\pm 0.4\%$}
 & $78.2\%$ \scriptsize{$\pm 0.5\%$}
 & $77.1\%$ \scriptsize{$\pm 0.6\%$}
 & $74.3\%$ \scriptsize{$\pm 0.4\%$}\\
 ABMLL (ours) 
 & $82.2\%$ \scriptsize{$\pm 0.1\%$} 
 & \bm{$80.6\%$} \scriptsize{$\pm 0.4\%$} 
 & \bm{$80.5\%$} \scriptsize{$\pm 0.5\%$} 
 & \bm{$80.8\%$} \scriptsize{$\pm 0.6\%$}
 & \bm{$79.0\%$} \scriptsize{$\pm 0.4\%$}\\
 \bottomrule
 \label{tab:a}
\end{tabular}}}
\resizebox{0.82\textwidth}{!}{
\subtable[Para.]{
\begin{tabular}{llllll} 
 \toprule
 Method & $0\%$ Pruned & $1\%$ Pruned & $10\%$ Pruned & $20\%$ Pruned & $30\%$ Pruned\\
 \midrule
 Pretrained 
 & $57.0\%$ \scriptsize{$\pm 0.0\%$}
 & $51.2\%$ \scriptsize{$\pm 0.0\%$}
 & $51.5\%$ \scriptsize{$\pm 0.0\%$}
 & $52.1\%$ \scriptsize{$\pm 0.1\%$}
 & $51.8\%$ \scriptsize{$\pm 0.2\%$}
 \\
 Regular LoRA 
 & $59.9\%$ \scriptsize{$\pm 0.4\%$}
 & $54.8\%$ \scriptsize{$\pm 0.2\%$}
 & $55.0\%$ \scriptsize{$\pm 0.1\%$}
 & $53.3\%$ \scriptsize{$\pm 0.1\%$}
 & $53.4\%$ \scriptsize{$\pm 0.3\%$}
 \\
 Struct. LoRA 
 & $59.9\%$ \scriptsize{$\pm 0.4\%$}
 & $56.0\%$ \scriptsize{$\pm 0.1\%$}
 & $55.5\%$ \scriptsize{$\pm 0.1\%$}
 & $55.1\%$ \scriptsize{$\pm 0.2\%$}
 & $52.9\%$ \scriptsize{$\pm 0.6\%$}
 \\
 Reptile 
 & \bm{$61.8\%$} \scriptsize{$\pm 0.2\%$} 
 & $56.1\%$ \scriptsize{$\pm 0.6\%$}
 & $56.0\%$ \scriptsize{$\pm 0.6\%$}
 & $54.8\%$ \scriptsize{$\pm 0.6\%$}
 & $53.2\%$ \scriptsize{$\pm 0.9\%$}
 \\
 ABMLL (ours) 
 & \bm{$61.6\%$} \scriptsize{$\pm 1.9\%$} 
 & \bm{$61.1\%$} \scriptsize{$\pm 1.0\%$} 
 & \bm{$61.0\%$} \scriptsize{$\pm 0.9\%$}
 & \bm{$60.1\%$} \scriptsize{$\pm 1.1\%$} 
 & \bm{$57.7\%$} \scriptsize{$\pm 1.6\%$}
 \\
 \bottomrule
 \label{tab:a}
\end{tabular}}}
\label{tab:prune}
\end{table}

\subsection{Ablation Studies}

Table 5 shows ABMLL performance across values of $\beta$ on cls-45. Results show that $\log_{10} \beta = 0$, i.e., $\beta=1$, drops performance significantly. Thus, it is critical to balance reconstruction error and the KL terms that control how close the task-specific parameters $\phi_i$ are to global parameters $\theta$, and how close $\theta$ are to the prior $p(\theta)$. At the other extreme where the KL terms are too heavily tempered, the objective becomes degenerate, leading to poor performance ($\log_{10} \beta = -16$).  In our experiments we searched for $\log_{10} \beta \in \{0,4,5,6,7,8,9,10\}$ with one random seed on cls-45 and identified the optimal value as $\log_{10} \beta = -8$. We use this setting for $\beta$ across all experiments.

In summary, while losses do not vary much with small changes in $\beta$, this study demonstrates that two extremes of $\beta$ drop performance significantly: either one where no balancing is used ($\beta=1$), or one that essentially drops the KL terms with an extremely small $\beta$. 

\begin{table}[t]
\centering
\caption{Validation accuracy and ECE across values of $\beta$ on cls-45. Lower $\beta$ means that the KL terms are more tempered, with pure maximization of data log likelihood in the limit.}
\resizebox{0.45\columnwidth}{!}{
\begin{tabular}{lcccc}
 \toprule
 $\log_{10}\beta$ & 0 & -5 & -8 & -16 \\
 \midrule
 Accuracy $\uparrow$ & $64.5\%$ & $70.4\%$ & \bm{$75.2\%$} & $60.6\%$ \\
 ECE $\downarrow$ & $0.395$ & $0.289$ & \bm{$0.262$} & $0.395$ \\
 \bottomrule
\end{tabular}}
\label{tab:beta}
\end{table}

\section{Discussion}

Our results show that a Bayesian approach to meta-learning for LoRA fine-tuning achieves strong generalization performance, measured in both accuracy and robustness, without a significant increase in computational cost. In the remainder of the paper we consider the implications of these results for broader questions about Bayesian methods and inductive biases for large language models. We also identify some of the limitations of our analyses and directions for future work.

\paragraph{Bridging the gap between Bayesian deep learning and LLMs.}
Bayesian modeling offers a principled way to encode prior knowledge and represent uncertainty \citep{Blei2014BuildCC, Griffiths2008}, and has been extensively studied in smaller neural networks \citep{MACKAY199573, blundell_2015}. For LLMs, however, the computational cost of Bayesian methods has limited their use, despite their promise for uncertainty quantification, interpretability, and domain adaptation \citep{pmlr-v235-papamarkou24b}. Our work helps bridge this gap by using LoRA to make Bayesian meta-learning practical for LLM fine-tuning.

\paragraph{Inductive bias in the era of LLMs.}
In-context learning \citep{min-etal-2022-metaicl, chen-etal-2022-meta} is a practical way to incorporate domain knowledge, but it can struggle when the relevant information is complex or unfamiliar \citep{dou2026clbenchbenchmarkcontextlearning}. Gradient-based meta-learning of model weights is often more expressive, but substantially more expensive. Data synthesis offers another route to injecting inductive bias \citep{mccoy_inductive}. Our results suggest a scalable alternative: in-weights meta-learning through Bayesian LoRA adaptation. This approach improves generalization, can be combined with in-context learning, and maintains memory overhead that is constant in the number of tasks.

\paragraph{Meta-learning as a potential post-training stage for LLMs.} We have shown that meta-learning on the cls45 split provides strong performance on unseen tasks that are not closely related. An open problem to explore further is finding a stronger, larger-scale general-purpose dataset for metalearning in order to produce stronger few-shot performance on various unseen domain applications. 




\paragraph{Conclusion.}

Meta-learning is an effective method for supporting better generalization across datasets, but its demands on computation and memory can make it difficult to apply to large language models. We have shown how meta-learning can be applied to LLMs at scale by combining Amortized Bayesian Meta-Learning with Low-Rank Adaptation. This approach results in higher accuracy and across several benchmarks, and produces models that are more robust to pruning than existing methods for meta-learning in LLMs. Furthermore, being able to use meta-learning in these models allows us to demonstrate that it is complementary to in-context learning, allowing in-context examples to be used more effectively.




\bibliography{colm2026_conference}
\bibliographystyle{colm2026_conference}

\newpage
\appendix
\section{Appendix}

\subsection{Algorithm}
\label{sec:appendix-alg}

Full algorithm for ABMLL is given in Algorithm \ref{alg:abmll}.

\begin{algorithm}[t]
   \caption{One epoch in the ABMLL algorithm. The ``test section'' does not need to be performed every epoch.}
   \label{alg:abmll}
   \textbf{Input:} Likelihood model $p(D_i | \phi_i)$, prior $p(\theta)$ and $p(\phi|\theta)$, variational posterior $q_{\theta}(\phi_i|D_i)$, with trainable parameters $\mathbf{B}, \mathbf{A}$; constant $c, \beta$; number of tasks $M$ and inner-loop size $K$. \\
   \begin{algorithmic}
   \STATE \textbf{Training section}
   \FOR{task $i \in \{1,2,...,M\}$}
   \STATE Inner-loop:
    \FOR{$k \in \{1,2,...,K\}$}
        \STATE Draw batch $D_i$ from task $i$ dataset.
        \STATE Run a step of gradient descent to minimize w.r.t. $\phi_i$,\\
        \begin{small}
        $-\mathbb{E}_{q_{\theta}(\phi_i|D_i)}[\log p(D_i|\phi_i)]+ \beta\text{KL}\big(q_{\theta}(\phi_i|D_i) \big| \big| p(\phi_i|\theta) \big).$
        \end{small}
    \ENDFOR
   \STATE Outer-loop: \\Run a step of gradient descent to minimize w.r.t. $\theta$,\\
   \begin{small}
        $-\mathbb{E}_{q_{\theta}(\phi_i|D_i)}[\log p(D_i|\phi_i)]+ \beta\text{KL}\big(q_{\theta}(\phi_i|D_i) \big| \big| p(\phi_i|\theta) \big)+ \beta\text{KL}(q(\theta) ||p(\theta)).$
   \end{small}
   \STATE $\mathbf{A}_{\mu_{\phi}} \leftarrow \mathbf{A}_{\mu_{\theta}}$, $\;\;\;\mathbf{A}_{\sigma_{\phi}} \leftarrow \mathbf{A}_{\sigma_{\theta}}$
   \STATE $\mathbf{B}_{\mu_{\phi}} \leftarrow \mathbf{B}_{\mu_{\theta}}$, $\;\;\;\mathbf{B}_{\sigma_{\phi}} \leftarrow \mathbf{B}_{\sigma_{\theta}}$
   \ENDFOR
   \end{algorithmic}
   \begin{algorithmic}
   \STATE \textbf{Test section}
   \STATE Take unseen task $i$. Create a copy of the above weights as $\theta^c, \phi_i^{c}$, and on the new weights:
    \FOR{$k \in \{1,2,...,K\}$}
        \STATE Draw batch $D_i$ from task $i$ dataset.
        \STATE Run a step gradient descent to minimize w.r.t $\phi_i^{c}$,\\
        \begin{small}
        $-\mathbb{E}_{q_{\theta^c}(\phi_i^{c}|D_i)}[\log p(D_i|\phi_i^{c})].$
        \end{small}
   \ENDFOR
   \STATE Evaluate on rest of data in task $i$.
   \STATE Delete the weights copy $\theta^c,\phi_i^{c}$ and reload the weights $\theta,\phi_i$.
   \end{algorithmic}
   \textbf{Output:} $\mathbf{B}$, $\mathbf{A}$.\\
   \label{alg:abmll}
\end{algorithm}

\subsection{Additional Experiment Results}
\label{appendix-exp-results}

Following Table \ref{tab:prune} in the main text, Table \ref{tab:prune-appendix} shows pruning results on the rest of datasets. Overall, ABMLL is significantly more robust against pruning than the other methods. MCQA dataset shows a few unique phenomena: ABMLL and Reptile are generally tied on robustness to pruning, and the pretrained model has better performance under small pruning than without pruning. A possible reason for the first phenomenon is that common-sense reasoning depends more on the pretrained capabilities, and learning generalization on this dataset is roughly equally difficult across methods.

\subsection{Experiment Details}
\label{sec:appendix-exp}
Here we detail the experimental setup used in our experiments. We use PyTorch with Torchtune to fine-tune \textsc{Llama3-8B-Chat}. Each experiment uses a single A100 GPU with 40GB memory. All methods use the AdamW optimizer \citep{loshchilov2017decoupled}, batch-size of 2, inner-loops with 5 gradient steps, LoRA adapters with rank $=8$ following the standard practice, and learning rate is tuned in $\small{[10^{-6}, 10^{-4}]}$. For ABMLL, $\small{\beta = 10^{-8}}$, $\small{c=e^{-20}}$. For the gamma prior, $\small{a_0=1, b_0 = 0.01}$ following \citet{ravi2018amortized}. 

We detail hyperparameters in Table \ref{tab:appendix-hyperparam}. All methods use the same LoRA setup, which is detailed in Table \ref{tab:appendix-lora}.

\subsection{Datasets}
\label{sec:appendix-data}

We use the setup of \citet{ye-etal-2021-crossfit} for our cls-45 and cls-23 datasets. We utilized the codebase provided by \citet{min-etal-2022-metaicl} to setup the datasets. Additionally, we focus on problems that can be converted into the multiple choice format. This allows us to evaluate the calibration error of models. Filtering for questions with at most four choices, we get the following training, validation, and test splits of these datasets.

\textbf{cls-45 training:} ['superglue-rte', 'tweet\_eval-sentiment', 'glue-rte', 'superglue-wsc', 'glue-mrpc', 'tweet\_eval-stance\_hillary', 'tweet\_eval-offensive', 'hatexplain', 'glue-cola', 'sick', 'paws', 'ethos-sexual\_orientation', 'glue-qqp', 'tweet\_eval-emotion', 'sms\_spam', 'health\_fact', 'glue-mnli', 'imdb', 'ethos-disability', 'glue-wnli', 'scitail', 'glue-sst2', 'tweet\_eval-stance\_abortion', 'tweet\_eval-stance\_climate', 'glue-qnli', 'ethos-directed\_vs\_generalized', 'ade\_corpus\_v2-classification', 'hate\_speech\_offensive', 'superglue-wic', 'google\_wellformed\_query', 'tweet\_eval-irony', 'ethos-gender', 'rotten\_tomatoes', 'kilt\_fever']

\textbf{cls-45 validation and testing:} ['tweet\_eval-stance\_feminist', 'ethos-national\_origin', 'tweet\_eval-hate', 'ag\_news', 'anli', 'hate\_speech18', 'poem\_sentiment', 'climate\_fever', 'medical\_questions\_pairs', 'tweet\_eval-stance\_atheism', 'ethos-race', 'ethos-religion', 'superglue-cb', 'wiki\_qa', 'yelp\_polarity']

\textbf{cls-23 training:} ['blimp-ellipsis\_n\_bar\_2','blimp-sentential\_negation\_npi\_scope',
 'crows\_pairs',
 'hellaswag',
 'openbookqa',
 'piqa',
 'quartz-no\_knowledge',
 'sciq',
 'ethos-disability',
 'ethos-sexual\_orientation',
 'glue-cola',
 'glue-mnli',
 'glue-mrpc',
 'glue-qqp',
 'glue-rte',
 'glue-wnli',
 'hatexplain',
 'health\_fact',
 'imdb',
 'paws',
 'sick',
 'sms\_spam',
 'superglue-rte',
 'superglue-wsc',
 'tweet\_eval-emotion',
 'tweet\_eval-offensive',
 'tweet\_eval-sentiment',
 'tweet\_eval-stance\_hillary']

\textbf{cls-23 validation testing:} same as cls-45 validation testing.

\textbf{NLI training:} ['ade\_corpus\_v2-classification',
 'ag\_news',
 'climate\_fever',
 'ethos-directed\_vs\_generalized',
 'ethos-disability',
 'ethos-gender',
 'ethos-national\_origin',
 'ethos-race',
 'ethos-religion',
 'ethos-sexual\_orientation',
 'glue-cola',
 'glue-mrpc',
 'glue-qqp',
 'glue-sst2',
 'google\_wellformed\_query',
 'hate\_speech18',
 'hate\_speech\_offensive',
 'hatexplain',
 'health\_fact',
 'imdb',
 'kilt\_fever',
 'medical\_questions\_pairs',
 'paws',
 'poem\_sentiment',
 'rotten\_tomatoes',
 'sick',
 'sms\_spam',
 'superglue-wic',
 'superglue-wsc',
 'tab\_fact',
 'tweet\_eval-emotion',
 'tweet\_eval-hate',
 'tweet\_eval-irony',
 'tweet\_eval-offensive',
 'tweet\_eval-sentiment',
 'tweet\_eval-stance\_abortion',
 'tweet\_eval-stance\_atheism',
 'tweet\_eval-stance\_climate',
 'tweet\_eval-stance\_feminist',
 'tweet\_eval-stance\_hillary',
 'wiki\_qa',
 'yelp\_polarity']

\textbf{NLI validation and testing:} ['sick', 'glue-mnli', 'glue-wnli', 'scitail', 'glue-rte', 'anli', 'superglue-cb', 'glue-qnli']

\textbf{para training:} ['ade\_corpus\_v2-classification',
 'ag\_news',
 'anli',
 'climate\_fever',
 'ethos-directed\_vs\_generalized',
 'ethos-disability',
 'ethos-gender',
 'ethos-national\_origin',
 'ethos-race',
 'ethos-religion',
 'ethos-sexual\_orientation',
 'glue-cola',
 'glue-mnli',
 'glue-qnli',
 'glue-rte',
 'glue-sst2',
 'glue-wnli',
 'google\_wellformed\_query',
 'hate\_speech18',
 'hate\_speech\_offensive',
 'hatexplain',
 'health\_fact',
 'imdb',
 'kilt\_fever',
 'poem\_sentiment',
 'rotten\_tomatoes',
 'scitail',
 'sick',
 'sms\_spam',
 'superglue-cb',
 'superglue-rte',
 'superglue-wic',
 'superglue-wsc',
 'tab\_fact',
 'tweet\_eval-emotion',
 'tweet\_eval-hate',
 'tweet\_eval-irony',
 'tweet\_eval-offensive',
 'tweet\_eval-sentiment',
 'tweet\_eval-stance\_abortion',
 'tweet\_eval-stance\_atheism',
 'tweet\_eval-stance\_climate',
 'tweet\_eval-stance\_feminist',
 'tweet\_eval-stance\_hillary',
 'wiki\_qa',
 'yelp\_polarity']

\textbf{para validation and testing:} ["glue-mrpc", "glue-qqp", "medical\_questions\_pairs", "paws"]

\textbf{MCQA training:} [
 'hate\_speech\_offensive',
 'google\_wellformed\_query',
 'glue-sst2',
 'scitail',
 'ag\_news',
 'art',
 'paws',
 'glue-qnli',
 'ade\_corpus\_v2-classification',
 'hatexplain',
 'glue-qqp',
 'kilt\_fever',
 'glue-mnli',
 'tab\_fact',
 'tweet\_eval-offensive',
 'imdb',
 'anli',
 'yelp\_polarity'
]

\textbf{MCQA validation and testing:} ['codah',
 'cosmos\_qa',
 'dream',
 'hellaswag',
 'openbookqa',
 'quarel',
 'quartz-no\_knowledge',
 'quartz-with\_knowledge',
 'sciq',
 'swag',
 'wino\_grande',
 'wiqa']

We also show an example of a question from Winogrande, demonstrating the format that we use across these datasets:

\begin{quote}
Return the label of the correct answer for the question below.\\

Question: Jason approached Steven to deliver the official subpoena and court summons, because \_ was being sued. \\

Choices:\\
A) Jason\\
B) Steven\\
\end{quote}

\begin{table}
\centering
\caption{Pruning results across methods on the rest of datasets. In each column except the first, a certain percentage of neurons in each layer embedding is set to zero. } 
\resizebox{0.9\textwidth}{!}{
\subtable[cls-45.]{
\begin{tabular}{llllll} 
 \toprule
 Method & $0\%$ Pruned & $1\%$ Pruned & $10\%$ Pruned & $20\%$ Pruned & $30\%$ Pruned\\
 \midrule
 Pretrained 
 & $26.1\%$ \scriptsize{$\pm 0.1\%$}
 & $24.4\%$ \scriptsize{$\pm 0.0\%$}
 & $23.0\%$ \scriptsize{$\pm 0.1\%$}
 & $18.8\%$ \scriptsize{$\pm 0.1\%$}
 & $13.5\%$ \scriptsize{$\pm 0.1\%$}
 \\
 Regular LoRA 
 & $71.6\%$ \scriptsize{$\pm 0.4\%$}
 & $65.8\%$ \scriptsize{$\pm 0.9\%$}
 & $65.6\%$ \scriptsize{$\pm 0.9\%$}
 & $65.1\%$ \scriptsize{$\pm 0.7\%$}
 & $61.9\%$ \scriptsize{$\pm 0.8\%$}
 \\
 Struct. LoRA 
 & $74.5\%$ \scriptsize{$\pm 0.4\%$}
 & $73.8\%$ \scriptsize{$\pm 0.3\%$}
 & $73.8\%$ \scriptsize{$\pm 0.3\%$}
 & $72.9\%$ \scriptsize{$\pm 0.3\%$}
 & $72.7\%$ \scriptsize{$\pm 0.2\%$}
 \\
 Reptile 
 & $73.0\%$ \scriptsize{$\pm 0.2\%$} 
 & $70.8\%$ \scriptsize{$\pm 0.4\%$}
 & $71.0\%$ \scriptsize{$\pm 0.3\%$}
 & $70.8\%$ \scriptsize{$\pm 0.2\%$}
 & $69.6\%$ \scriptsize{$\pm 0.5\%$}
 \\
 ABMLL (ours) 
 & \bm{$75.2\%$} \scriptsize{$\pm 1.9\%$} 
 & \bm{$75.6\%$} \scriptsize{$\pm 0.2\%$} 
 & \bm{$75.2\%$} \scriptsize{$\pm 0.5\%$}
 & \bm{$75.3\%$} \scriptsize{$\pm 0.0\%$} 
 & \bm{$74.3\%$} \scriptsize{$\pm 0.2\%$}
 \\
 \bottomrule
 \label{tab:a}
\end{tabular}}}
\resizebox{0.9\textwidth}{!}{
\subtable[cls-23.]{
\begin{tabular}{llllll} 
 \toprule
 Method & $0\%$ Pruned & $1\%$ Pruned & $10\%$ Pruned & $20\%$ Pruned & $30\%$ Pruned\\
 \midrule
 Pretrained 
 & $26.0\%$ \scriptsize{$\pm 0.1\%$}
 & $24.4\%$ \scriptsize{$\pm 0.0\%$}
 & $22.9\%$ \scriptsize{$\pm 0.1\%$}
 & $18.8\%$ \scriptsize{$\pm 0.1\%$}
 & $13.5\%$ \scriptsize{$\pm 0.1\%$}
 \\
 Regular LoRA 
 & $71.4\%$ \scriptsize{$\pm 0.4\%$}
 & $64.5\%$ \scriptsize{$\pm 0.5\%$}
 & $64.1\%$ \scriptsize{$\pm 0.5\%$}
 & $63.5\%$ \scriptsize{$\pm 0.6\%$}
 & $60.5\%$ \scriptsize{$\pm 0.6\%$}
 \\
 Struct. LoRA 
 & $71.4\%$ \scriptsize{$\pm 0.4\%$}
 & $71.5\%$ \scriptsize{$\pm 0.3\%$}
 & $71.6\%$ \scriptsize{$\pm 0.3\%$}
 & $71.0\%$ \scriptsize{$\pm 0.2\%$}
 & $70.1\%$ \scriptsize{$\pm 0.4\%$}
 \\
 Reptile 
 & $72.7\%$ \scriptsize{$\pm 0.2\%$} 
 & $71.6\%$ \scriptsize{$\pm 0.2\%$}
 & $71.6\%$ \scriptsize{$\pm 0.1\%$}
 & $71.3\%$ \scriptsize{$\pm 0.1\%$}
 & $70.5\%$ \scriptsize{$\pm 0.3\%$}
 \\
 ABMLL (ours) 
 & \bm{$73.3\%$} \scriptsize{$\pm 1.9\%$} 
 & \bm{$72.9\%$} \scriptsize{$\pm 0.5\%$} 
 & \bm{$73.0\%$} \scriptsize{$\pm 0.4\%$}
 & \bm{$72.5\%$} \scriptsize{$\pm 0.3\%$} 
 & \bm{$71.5\%$} \scriptsize{$\pm 0.4\%$}
 \\
 \bottomrule
 \label{tab:a}
\end{tabular}}}
\resizebox{0.9\textwidth}{!}{
\subtable[MCQA.]{
\begin{tabular}{llllll} 
 \toprule
 Method & $0\%$ Pruned & $1\%$ Pruned & $10\%$ Pruned & $20\%$ Pruned & $30\%$ Pruned\\
 \midrule
 Pretrained 
 & $71.8\%$ \scriptsize{$\pm 0.1\%$}
 & $73.0\%$ \scriptsize{$\pm 0.0\%$}
 & $72.8\%$ \scriptsize{$\pm 0.0\%$}
 & $71.4\%$ \scriptsize{$\pm 0.1\%$}
 & $64.7\%$ \scriptsize{$\pm 0.2\%$}
 \\
 Regular LoRA 
 & $76.1\%$ \scriptsize{$\pm 0.1\%$}
 & $74.2\%$ \scriptsize{$\pm 0.4\%$}
 & $73.8\%$ \scriptsize{$\pm 0.4\%$}
 & $72.4\%$ \scriptsize{$\pm 0.4\%$}
 & $70.7\%$ \scriptsize{$\pm 0.3\%$}
 \\
 Struct. LoRA 
 & $76.0\%$ \scriptsize{$\pm 0.0\%$}
 & $74.6\%$ \scriptsize{$\pm 0.0\%$}
 & $74.4\%$ \scriptsize{$\pm 0.1\%$}
 & $72.4\%$ \scriptsize{$\pm 0.0\%$}
 & $70.9\%$ \scriptsize{$\pm 0.1\%$}
 \\
 Reptile 
 & \bm{$77.7\%$} \scriptsize{$\pm 0.1\%$} 
 & \bm{$76.2\%$} \scriptsize{$\pm 0.3\%$}
 & \bm{$76.1\%$} \scriptsize{$\pm 0.4\%$}
 & \bm{$75.0\%$} \scriptsize{$\pm 0.3\%$}
 & \bm{$73.3\%$} \scriptsize{$\pm 0.2\%$}
 \\
 ABMLL (ours) 
 & $77.4\%$ \scriptsize{$\pm 0.1\%$} 
 & \bm{$76.2\%$} \scriptsize{$\pm 0.4\%$} 
 & \bm{$76.1\%$} \scriptsize{$\pm 0.5\%$}
 & \bm{$75.4\%$} \scriptsize{$\pm 0.4\%$} 
 & \bm{$73.3\%$} \scriptsize{$\pm 0.5\%$}
 \\
 \bottomrule
 \label{tab:a}
\end{tabular}}}
\label{tab:prune-appendix}
\end{table}

\begin{table}
\centering
\caption{Hyperparameters for experiments.}
\resizebox{0.87\columnwidth}{!}
{\begin{tabular}{lllll} 
 \toprule
 Hyperparameter & ABMLL & Structured LoRA & Regular LoRA & Reptile \\
 \midrule
 Inner loop / regular learning rate & $10^{-5}$ & $10^{-5}$ & $5\cdot10^{-5}$ & $10^{-5}$ \\
 Outer loop learning rate & $5\cdot 10^{-5}$ & NA & NA & NA \\
 Test adaptation learning rate & $10^{-5}$ & $10^{-5}$ & $10^{-5}$ & $10^{-5}$ \\
 Step size $\epsilon$ & NA & NA & NA & $0.5$ \\
 \bottomrule
\end{tabular}}
\label{tab:appendix-hyperparam}
\end{table}

\begin{table}
\centering
\caption{LoRA setup.}
\resizebox{0.7\columnwidth}{!}
{\begin{tabular}{ll} 
 \toprule
 LoRA rank & $8$ \\
 LoRA $\alpha$ & $16$ \\
 Modules using LoRA & Q Projection, V Projection, Output Projection \\
 \bottomrule
\end{tabular}}
\label{tab:appendix-lora}
\end{table}

\end{document}